\documentclass[final]{cvpr}

\usepackage{times}
\usepackage{epsfig}
\usepackage{amsmath,amssymb} 
\usepackage{color}
\usepackage{cite}
\usepackage{epsfig}
\usepackage{graphicx}
\usepackage{subfig}

\usepackage{cite}
\usepackage{comment}
\usepackage{adjustbox}
\usepackage{textcomp}
\usepackage{array,multirow,graphicx}
\usepackage[table]{xcolor}
\usepackage{colortbl}
\usepackage{bbm}
\usepackage{makecell}
\usepackage[percent]{overpic}

\usepackage{booktabs}
\usepackage{arydshln}
\usepackage{url}

\usepackage{pgfplots}
\usepackage{filecontents}
\usepackage{wrapfig}

\usepackage{tikz}
\usetikzlibrary{arrows}
\usetikzlibrary{calc} 

\usepackage{lipsum}

\usepackage[pagebackref=true,breaklinks=true,colorlinks,bookmarks=false]{hyperref}

\def\best{\bf \cellcolor[gray]{0.85}}
\def\secbest{\cellcolor[gray]{0.92}}

\definecolor{TableRed}{HTML}{800000}
\newcommand{\TextRed}[1]{\textcolor{TableRed}{#1}}

\def\cvprPaperID{3890} 
\def\confYear{CVPR 2021}

\begin{document}

\title{ResNeSt: Split-Attention Networks}

\author{
Hang Zhang$^1$, Chongruo Wu$^2$, Zhongyue Zhang$^3$, Yi Zhu$^4$,
Haibin Lin$^5$, Zhi Zhang$^4$, \\
Yue Sun$^6$, Tong He$^4$, Jonas Mueller$^4$,
R. Manmatha$^4$, Mu Li$^4$, Alexander Smola$^4$ \\\\
Facebook$^1$, UC Davis$^2$, Snap$^3$, Amazon$^4$, ByteDance$^5$, SenseTime$^6$ \\
{\tt\small zhanghang@fb.com, crwu@ucdavis.edu, zzhang5@snapchat.com, haibin.lin@bytedance.com,} \\ 
{\tt\small sunyue1@sensetime.com, \{yzaws,zhiz,htong,jonasmue,manmatha,mli,smola\}@amazon.com}
}

\maketitle

\begin{abstract}

It is well known that featuremap attention  and multi-path representation are important for visual recognition. 
In this paper, we present a modularized architecture, which applies the channel-wise attention on different network branches to leverage their success in capturing cross-feature interactions and learning diverse representations. 
Our design results in a simple and unified computation block, which can be parameterized using only a few variables. 
Our model, named ResNeSt, outperforms EfficientNet in accuracy and latency trade-off on image classification. 
In addition, ResNeSt has achieved superior transfer learning results on several public benchmarks serving as the backbone, and has been adopted by the winning entries of COCO-LVIS challenge. 
The source code for complete system and pretrained models are publicly available.

\end{abstract}

\section{Introduction}

Deep convolutional neural networks (CNNs) have become 
the fundamental 
approach for image classification and other transfer learning tasks in computer vision. 
As the key component of the CNNs, a convolutional layer learns a set of filters which aggregates the neighborhood information with spatial and channel connections. 
This operation is suitable to capture {\it correlated features} with the output channels densely connected to each input channel. 
Inception models~\cite{szegedy2015going,szegedy2016rethinking} explore the multi-path representation to learn {\it independent features}, where the input is split into a few lower dimensional embeddings, transformed by different sets of convolutional filters and then merged by concatenation. 
This strategy encourages the feature exploration by decoupling the input channel connections~\cite{xie2016aggregated}. 









\def\mystrut{\vphantom{hg}}
\pgfplotsset{
    legend image with text/.style={
        legend image code/.code={%
            \node[anchor=center] at (0.3cm,0cm) {#1};
        }
    },
}

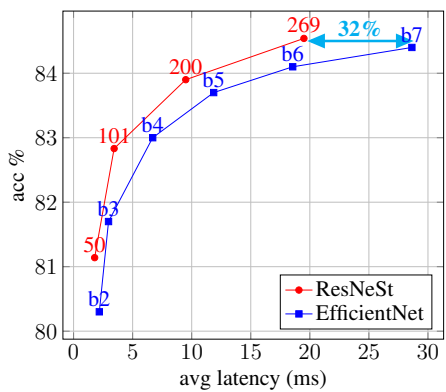
\begin{figure}[t]
\begin{center}
 \hspace{-2em}
    \scalebox{0.6}
      {
        \pgfplotsset{compat=1.3}
        \begin{tikzpicture}
            \begin{axis} [
                legend style={
                    font=\mystrut,
                    legend cell align=left,
                },
                legend style={font=\fontsize{14}{6}\selectfont},
                style={font=\fontsize{14}{6}\selectfont},
                legend pos=south east,
                scale only axis,
                ylabel={acc \%},
                xlabel={avg latency (ms)},
                grid=both,
            ]
                \addplot[
                    color=red, 
                    mark=*, 
                    nodes near coords,
                    point meta=explicit symbolic] table [x=a, y=b, meta=c, col sep=comma] {resnest.csv};
                \addlegendentry{ResNeSt}
                
                \addplot [
                    color=blue, 
                    mark=square*, 
                    nodes near coords,
                    point meta=explicit symbolic
                    ] table [x=a, y=b, meta=c, col sep=comma] {efficientnet.csv};
                \addlegendentry{EfficientNet}
                
                
                
            \node (O) at (225,440) {\textcolor{cyan}{\bf 32\%}};
            \draw[draw=cyan,>=triangle 45,ultra thick, <->] (180,420) -- (270,420);
            \end{axis}
        \end{tikzpicture}
    }
\end{center}
\caption{
ResNeSt outperforms EfficientNet in accuracy-latency trade-offs on GPU. Notably, ResNeSt-269 has achieved better accuracy than EfficientNet-B7 with 32\% less latency. (details in Section~\ref{sec:exp}). 
}
\label{tab:abstract}
\end{figure}

The neuron connections in visual cortex have inspired the development of CNNs in the past decades~\cite{hubel1962receptive}. 
The main theme of visual representation learning is discovering salient features for a given task~\cite{zhou2014learning}. 
Prior work has modeled spatial and channel dependencies~\cite{bell2016inside,newell2016stacked,hu2017squeeze}, and incorporated attention mechanism~\cite{hu2017squeeze,sknet,Wang_2018_CVPR}. 
SE-like channel-wise attention~\cite{hu2017squeeze} employs global pooling to squeeze the channel statistics, and predicts a set of attention factors to apply channel-wise multiplication with the original featuremaps. 
This mechanism models the interdependencies of featuremap channels, which uses the global context information to selectively highlight or de-emphasize the features~\cite{hu2017squeeze,sknet}. 
This attention mechanism is similar to attentional selection stage of human primary visual cortex~\cite{zhaoping2014understanding}, which finds the informative parts for recognizing objects. 
Human/animals perceive various visual patterns using the cortex in separate regions that respond to different and particular visual features~\cite{rajalingham2019reversible}. 
This strategy makes it easy to identify subtle but dominant differences of similar objects in the neural perception system. 
Similarly, if we can build a CNN architecture to capture individual salient attributes for different visual features, we would improve the network representation for image classification.

In this paper, we present a simple architecture which combines the channel-wise attention strategy with multi-path network layout. Our method captures cross-channel feature correlations, while preserving independent representation in the meta structure. 
A module of our network performs a set of transformations on low dimensional embeddings and concatenates their outputs as in a multi-path network. 
Each transformation incorporates channel-wise attention strategy to capture interdependencies of the featuremap. 
We further simplify the architecture to make each transformation share the same topology ({\it e.g.} Fig~\ref{fig:block} (Right)). We can parameterize the network architecture with only a few variables. In addition, such setting also allows us to accelerate the training using identical implementation with unified CNN operators. 
We refer to such computation block as {\it Split-Attention Block}. 
Stacking several Split-Attention blocks in ResNet style, we create a new ResNet variant which we refer to as \emph{ Split-Attention Network (ResNeSt)}.



We benchmark the performance of the proposed ResNeSt networks on ImageNet dataset~\cite{imagenet}. 
The proposed ResNeSt achieves better speed-accuracy trade-offs than state-of-the-art CNN models produced via neural architecture search~\cite{efficientnet} as shown in Table~\ref{tab:cls_sota}. 
In addition, we also study the transfer learning results on object detection, instance segmentation and semantic segmentation. 
The proposed ResNeSt has achieved superior performance on several gold-standard benchmarks when serving as the backbone network. 
For example, our Cascade-RCNN~\cite{cai2019cascade} model with ResNeSt-101 backbone achieves 48.3\% box mAP and 41.56\% mask mAP on MS-COCO instance segmentation.  
Our DeepLabV3~\cite{chen2017rethinking} model, again using a ResNeSt-101 backbone, achieves  mIoU of 46.9\% on the ADE20K scene parsing validation set, which surpasses the previous best result by more than 1\% mIoU.
Furthermore, ResNeSt has been adopted by the winning entries of 2020 COCO-LVIS challenge~\cite{wang2020seesaw,tan20201st,teamSSL}. 


\section{Related Work}


\paragraph{\bf CNN Architectures. }
Since AlexNet~\cite{krizhevsky2012imagenet}, deep convolutional neural networks~\cite{lecun1998gradient} have dominated image classification.  
With this trend, research has shifted from engineering handcrafted features to engineering network architectures. 
NIN~\cite{lin2013network} 
first uses a global average pooling layer to replace the heavy fully connected layers, and adopts $1\times 1$ convolutional layers to learn non-linear combination of the featuremap channels, which is the first kind of featuremap attention mechanism. 
VGG-Net~\cite{simonyan2014very} proposes a modular network design strategy, stacking the same type of network blocks repeatedly, which simplifies both the workflow of network design and transfer learning for downstream applications. 
Highway network~\cite{srivastava2015highway} introduces highway connections which makes the information flow across several layers without attenuation and helps the network convergence. 
Built on the success of the pioneering work, ResNet~\cite{he2015deep} introduces an identity skip connection which alleviates the difficulty of vanishing gradient in deep neural network and allows network to learn improved feature representations. 
ResNet has become one of the most successful CNN architectures which has been adopted in various computer vision applications. 
\vspace{-1em}

\paragraph{\bf Multi-path and featuremap Attention. }
Multi-path representation has shown success in GoogleNet~\cite{szegedy2015going}, in which each network block consists of different convolutional kernels. 
ResNeXt~\cite{xie2017aggregated} adopts group convolution~\cite{krizhevsky2012imagenet} in the ResNet bottle block, which converts the multi-path structure into a unified operation. 
SE-Net~\cite{hu2017squeeze} introduces a channel-attention mechanism by adaptively recalibrating the channel feature responses. 
Recently, SK-Net~\cite{sknet} brings the featuremap attention across two network branches. 
Inspired by the previous methods, our network integrates the channel-wise attention with multi-path network representation. 
\vspace{-1em}

\paragraph{\bf Neural Architecture Search. }
With increasing computational power, research interest has begun shifting from manually designed architectures to systematically searched architectures. 
Recent work explored efficient neural architecture search via parameter sharing~\cite{liu2018darts,pham2018efficient} and have achieved great success in low-latency and low-complexity CNN models~\cite{wu2019fbnet,cai2019once}. 
However, searching a large-scale neural network is still challenging due to the high GPU memory usage via parameter sharing with other architectures. 
EfficientNet~\cite{efficientnet} first searches in a small setting and then scale up the network complexity systematically. 
Instead, we build our model with ResNet meta architecture to scale up the network to deeper versions (from 50 to 269 layers). 
Our approach also augments the search spaces for neural architecture search and potentially improve the overall performance, which can be studied in the future work.

\section{Split-Attention Networks}

We now introduce the Split-Attention block, which enables featuremap attention across different featuremap groups in Section \ref{sec:splitatten}. 
Later, we describe our network instantiation and how to accelerate this architecture via standard CNN operators in Section \ref{sec:radix}.

\subsection{Split-Attention Block}
\label{sec:splitatten}

\begin{figure*}[t]
    \centering
    \includegraphics[width=\linewidth]{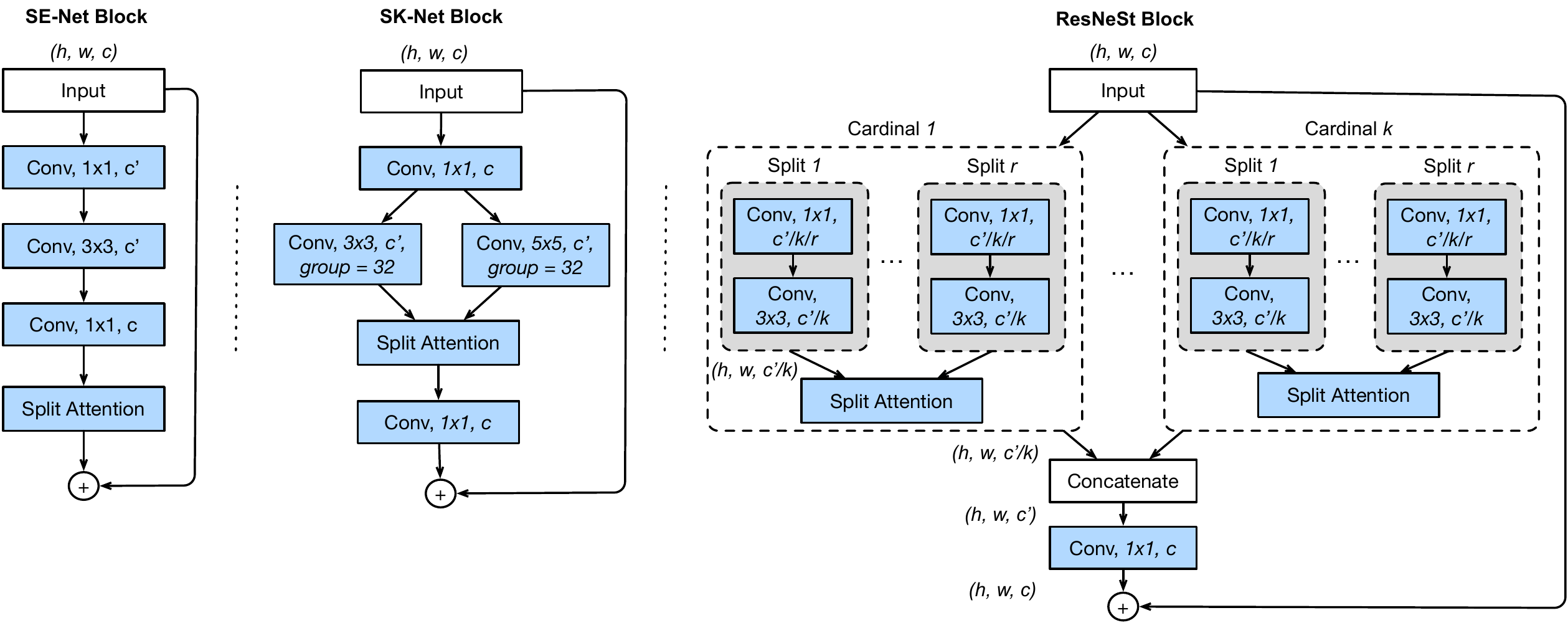}
    \caption{
    Comparing our ResNeSt block with SE-Net~\cite{hu2018squeeze} and SK-Net~\cite{sknet}. A detailed view of Split-Attention unit is shown in Figure~\ref{fig:splatunit}. 
    For simplicity, we show ResNeSt block in cardinality-major view (the featuremap groups with same cardinal group index reside next to each other). We use radix-major in the real implementation, which can be modularized and accelerated by group convolution and standard CNN layers (see supplementary material).  
    }
    \label{fig:block}
    \vspace{-1em}
\end{figure*}

Our {\it Split-Attention} block is a computational unit, consisting of {\it featuremap group} and {\it split attention} operations. Figure~\ref{fig:block} (Right) depicts an overview of a Split-Attention Block. 

\noindent {\bf Featuremap Group. } 
As in ResNeXt blocks~\cite{xie2017aggregated}, the feature can be divided into several groups
, and the number of featuremap groups is given by a \emph{cardinality} hyperparameter $K$. 
We refer to the resulting featuremap groups as \emph{cardinal groups}. 
In this paper, we introduce a new \emph{radix} hyperparameter $R$ that indicates the number of splits within a cardinal group, so the total number of feature groups is $G=K R$. 
We may apply a series of transformations $\{\mathcal{F}_1, \mathcal{F}_2, ... \mathcal{F}_G\}$ to each individual group, then the intermediate representation of each group is $U_i = \mathcal{F}_i(X) \text{, for } i \in \{1, 2, ...G\}$.

\begin{figure}
    \centering
    \includegraphics[width=0.65\linewidth]{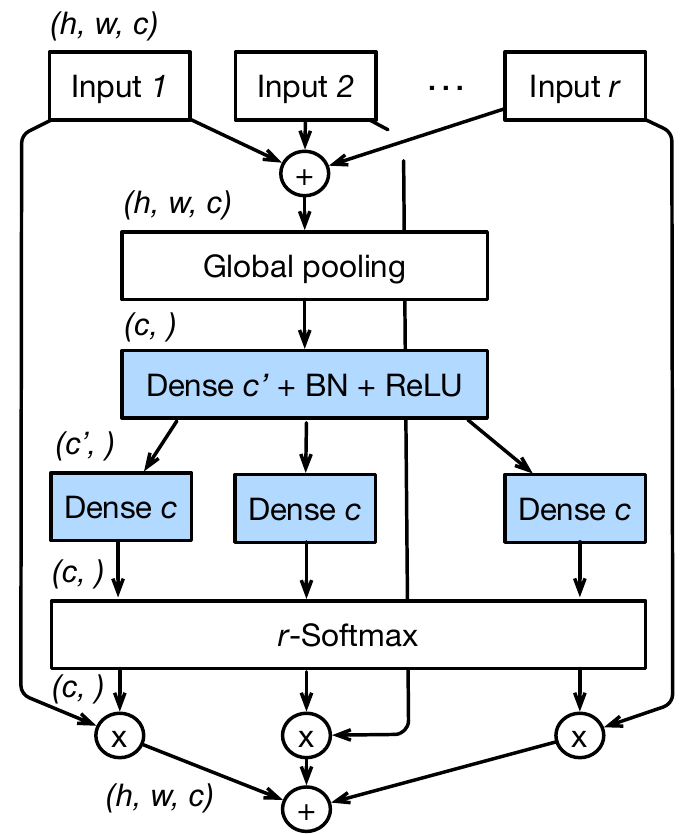}
    \caption{Split-Attention within a cardinal group. For easy visualization in the figure, we use $c=C/K$ in this figure.} 
    \label{fig:splatunit}
\end{figure}

\noindent  {\bf Split Attention in Cardinal Groups. } 
Following~\cite{hu2018squeeze,sknet}, a combined representation for each cardinal group can be obtained by fusing via an element-wise summation across multiple splits. The representation for $k$-th cardinal group is 
$\hat{U}^k = \sum_{j=R(k-1)+1}^{R k} U_j $, where $\hat{U}^k \in \mathbb{R}^{H\times W\times C/K}$ for $k\in{1,2,...K}$, and $H$, $W$ and $C$ are the block output featuremap sizes. 
Global contextual information with embedded channel-wise statistics can be gathered with global average pooling across spatial dimensions  $s^k\in\mathbb{R}^{C/K}$~\cite{sknet,hu2017squeeze}. Here the $c$-th component is calculated as:

\begin{equation}
    s^k_c = \frac{1}{H\times W} \sum_{i=1}^H\sum_{j=1}^W \hat{U}^k_c(i, j).
\end{equation}

A weighted fusion of the cardinal group representation $V^k\in\mathbb{R}^{H\times W\times C/K}$ is aggregated using channel-wise soft attention, where each featuremap channel is produced using a weighted combination over splits. Then the $c$-th channel is calculated as:
\begin{equation}
    V^k_c=\sum_{i=1}^R a^k_i(c) U_{R(k-1)+i} ,
\end{equation}
where $a_i^k(c)$ denotes a (soft) assignment weight given by:
\begin{equation}
a_i^k(c) =
\begin{cases}
  \frac{exp(\mathcal{G}^c_i(s^k))}{\sum_{j=1}^R exp(\mathcal{G}^c_j(s^k))} & \quad\textrm{if } R>1, \\
   \frac{1}{1+exp(-\mathcal{G}^c_i(s^k))} & \quad\textrm{if } R=1,\\
\end{cases}
\label{eq:radix}
\end{equation}
and mapping $\mathcal{G}_i^c$ determines the weight of each split for the $c$-th channel based on the global context representation $s^k$. 

\noindent{\bf ResNeSt Block. }
The cardinal group representations are then concatenated along the channel dimension: $V = Concat\{V^1,V^2,...V^K\}$.  
As in standard residual blocks, the final output $Y$ of our Split-Attention block is produced using a shortcut connection: $Y=V+X$, if the input and output featuremap share the same shape. 
For blocks with a stride, an appropriate transformation $\mathcal{T}$ is applied to the shortcut connection to align the output shapes:  $Y=V+\mathcal{T}(X)$. For example, $\mathcal{T}$ can be strided convolution or combined convolution-with-pooling.

\noindent{\bf Instantiation and Computational Costs.}
Figure~\ref{fig:block} (right) shows an instantiation of our Split-Attention block, in which the group transformation $\mathcal{F}_i$ is a $1\times 1$ convolution followed by a 
$3\times 3$ convolution, and the attention weight function $\mathcal{G}$ is parameterized using two fully connected layers with ReLU activation. 
The number of parameters and FLOPS of a Split-Attention block are roughly the same as a standard residual block~\cite{he2015deep,xie2016aggregated} with the same cardinality and number of channels.

\noindent{\bf Relation to Existing Attention Methods.}
First introduced in SE-Net~\cite{hu2017squeeze}, the idea of squeeze-and-attention (called \emph{excitation} in the original paper)  is to employ a global context to predict channel-wise attention factors. 
With $\text{radix} =1$, our Split-Attention block is applying a squeeze-and-attention operation to each cardinal group, while the SE-Net operates on top of the entire block regardless of multiple groups.  
SK-Net~\cite{sknet} introduces feature attention between two network streams. Setting $\text{radix} =2$, the Split-Attention block applies SK-like attention to each cardinal group. 
Our method generalizes prior work of featuremap attention~\cite{hu2017squeeze,sknet} within a cardinal group setting~\cite{xie2016aggregated}, and its implementation remains computationally efficient.
Figure~\ref{fig:block} shows an overall comparison with SE-Net and SK-Net blocks.

\subsection{Efficient Radix-major Implementation}
\label{sec:radix}

\begin{figure}[t]
    \centering
    \includegraphics[width=0.9\linewidth]{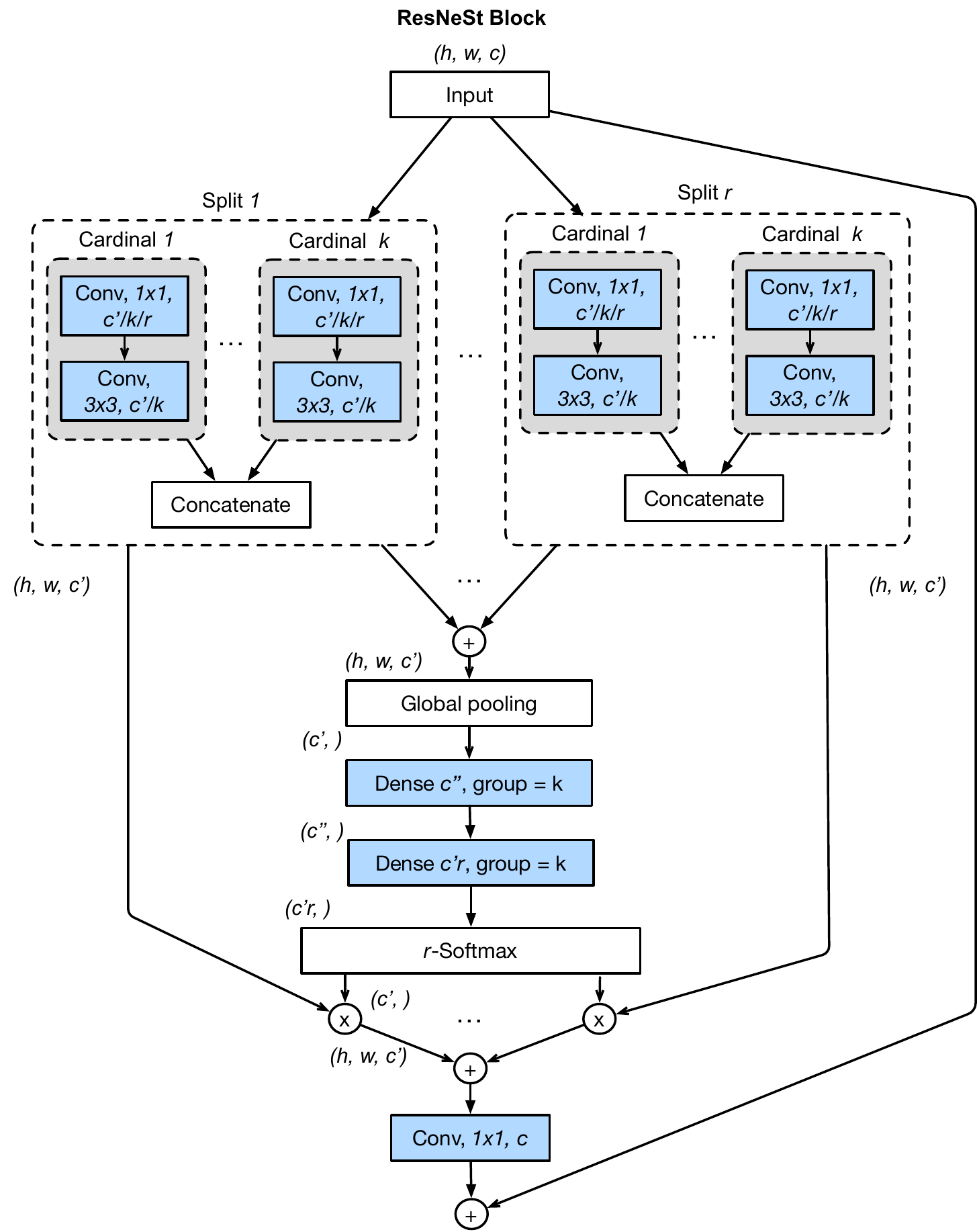}
    \caption{Radix-major implementation of ResNeSt block, where the featuremap groups with same radix index but different cardinality are next to each other physically. 
    This implementation can be implemented using unified CNN operators. (See details in Section~\ref{sec:radix}.)
    }
    \label{fig:raidx_major}
\end{figure}

We refer to the layout described in the previous section as  {\it cardinality-major implementation}, where the featuremap groups with the same cardinal index reside next to each other physically (Figure~\ref{fig:block} (Right)). 
The cardinality-major implementation is straightforward and intuitive, but is difficult to modularize and accelerate using standard CNN operators. 
For this, we introduce an equivalent {\it radix-major implementation}.

Figure~\ref{fig:raidx_major} gives an overview of the Split-Attention block in radix-major layout. 
The input featuremap is first divided into $R K$ groups, in which each group has a cardinality-index and radix-index. 
In this layout, the groups with same radix-index reside next to each other. 
Then, we can conduct a summation across different splits, so that the featuremap groups with the same cardinality-index but different radix-index are fused together. 
A global pooling layer aggregates over the spatial dimension, while keeps the channel dimension separated, which is identical to conducting global pooling to each individual cardinal groups then concatenate the results. 
Then two consecutive fully connected (FC) layers with number of groups equal to cardinality are added after pooling layer to predict the attention weights for each splits. 
The use of grouped FC layers makes it identical to apply each pair of FCs separately on top each cardinal groups. 

With this implementation, the first $1\times 1$ convolutional layers can be unified into one layer and the $3\times 3$ convolutional layers can be implemented using a single grouped convolution with the number of groups of $RK$. 
Therefore, the Split-Attention block is modularized using standard CNN operators.

\section{Network and Training}

We now describe the network design and training strategies used in our experiments. First, we detail a couple of tweaks that further improve performance, some of which have been empirically validated in~\cite{Xie2018bags}.

\subsection{Network Tweaks}

\noindent{\bf Average Downsampling. }
For transfer learning on dense prediction tasks such as detection or  segmentation, it becomes essential to preserve spatial information. 
Recent ResNet implementations usually apply the strided convolution at the $3\times 3$ layer instead of the previous $1\times 1$ layer to better preserve such information~\cite{he2019bag,hu2018squeeze}. 
Convolutional layers require handling featuremap boundaries with zero-padding strategies, which is often suboptimal when transferring to other dense prediction tasks.  
Instead of using strided convolution at the transitioning block (in which the spatial resolution is downsampled), we use an average pooling layer with a kernel size of $3\times 3$ .

\noindent{\bf Tweaks from ResNet-D.}
We also adopt two simple yet effective ResNet modifications introduced by~\cite{he2019bag}: 
(1) The first $7\times 7$ convolutional layer is replaced with three consecutive $3\times 3$ convolutional layers, which have the same receptive field size with a similar computation cost as the original design. (2) A $2\times2$ average pooling layer is added to the shortcut connection prior to the $1\times1$ convolutional layer for the transitioning blocks with stride of two.

\subsection{Training Strategy}

\noindent{\bf Large Mini-batch Distributed Training.}\footnote{Note that large mini-batch training does not improve network accuracy. Instead, it often degrades the results.}
For effectively training deep CNN models, we follow the prior work ~\cite{goyal2017accurate,li2017scaling,lin2019dynamic} to train our models using 8 servers (64 GPUs in total) in parallel. 
Our learning rates are adjusted according to a cosine schedule~\cite{huang2016densely, he2019bag}. 
We follow the common practice using linearly scaling-up the initial learning rate based on the mini-batch size. 
The initial learning rate is given by $\eta=\frac{B}{256} \eta_{base} $, where $B$ is the mini-batch size and we use $\eta_{base}=0.1$ as the base learning rate. 
This warm-up strategy is applied over the first 5 epochs, gradually increasing the learning rate linearly from 0 to the initial value for the cosine schedule~\cite{goyal2017accurate,lin2019dynamic}.
The batch normalization (BN) parameter $\gamma$ is initialized to zero in the final BN operation of each block, as has been suggested for large batch training~\cite{goyal2017accurate}.

\noindent{\bf Label Smoothing. }
Label smoothing was first used to improve the training of Inception-V2 ~\cite{szegedy2016rethinking}. 
Recall the cross entropy loss incurred by our network's predicted class probabilities $q$ is computed against ground-truth  $p$ as:
\begin{equation}
    \ell (p, q) = - \sum _{i=1}^K p_i \log q_i ,
\end{equation}
where $K$ is total number of classes, $p_i$ is the ground truth probability of the $i$-th class, and $q_i$ is the network's predicted probability for the $i$-th class. 
As in standard image classification,  $q_i =\frac{exp(z_i)}{\sum_{j=1}^K exp(z_j)}$ where $z_i$ are the logits produced by our network's ouput layer. When the provided labels are classes rather than class-probabilities (hard labels), $p_i=1$ if $i$ equals the ground truth class $c$, and is otherwise $=0$. Thus in this setting: $\ell_{hard}(p, q)=-\log q_c=-z_c+\log (\sum_{j=1}^K exp(z_j))$. During the final phase of training, the logits $z_j$ tend to be very small for $j\ne c$, while $z_c$ is being pushed to its optimal value $\infty$, and this can induce  overfitting~\cite{he2019bag,szegedy2016rethinking}.  
Rather than assigning hard labels as targets, label smoothing uses a smoothed ground truth probability:
\begin{equation}
p_i =
\begin{cases}
  1-\varepsilon & \quad\textrm{if } i = c, \\
   \varepsilon / (K-1) & \quad\textrm{otherwise}\\
\end{cases}
\end{equation}
with small constant $\varepsilon > 0$. 
This mitigates network overconfidence and overfitting.


\noindent{\bf Auto Augmentation. }
Auto-Augment~\cite{cubuk2019autoaugment} is a strategy that augments the training data with transformed images, where the transformations are learned adaptively. 
16 different types of image jittering transformations are introduced, and from these, one augments the data based on 24 different combinations of two consecutive transformations such as shift, rotation, and color jittering. 
The magnitude of each  transformation can be controlled with a relative parameter (e.g.\  rotation angle), and transformations may be probabilistically skipped.  

\noindent{\bf Mixup Training. }
Mixup is another data augmentation strategy that generates a weighted combinations of random image pairs from the training data~\cite{zhang2017mixup}. Given two images and their ground truth labels: $(x^{(i)}, y^{(i)}), (x^{(j)}, y^{(j)})$, a synthetic training example $(\hat{x}, \hat{y})$ is generated as:
\begin{eqnarray}
    \hat{x}&=\lambda x^i + (1-\lambda)x^j , \\
    \hat{y}&=\lambda y^i + (1-\lambda)y^j ,
\end{eqnarray}
where $\lambda \sim \text{Beta}(\alpha = 0.2)$ is independently sampled for each augmented example. 

\noindent{\bf Large Crop Size. }
Image classification research typically compares the performance of different networks operating on images that share the same crop size. 
ResNet variants~\cite{he2015deep,hu2017squeeze,xie2016aggregated,he2019bag} usually use a fixed training crop size of 224, while the Inception-Net  family~\cite{szegedy2015going,szegedy2017inception,szegedy2016rethinking} uses a training crop size of 299. 
Recently, the EfficientNet method~\cite{efficientnet} has demonstrated that increasing the input image size for a deeper and wider network may better trade off accuracy vs.\ FLOPS. For fair comparison, we use a crop size of 224 when comparing our ResNeSt with ResNet variants, and a crop size of 256 when comparing with other approaches.

\noindent{\bf Regularization. }
Very deep neural networks tend to overfit even for large  datasets~\cite{zhang2017polynet}. 
To prevent this, dropout regularization randomly masks out some neurons during training (but not during inference) to form an implicit network ensemble~\cite{srivastava2014dropout, hu2017squeeze,zhang2017polynet}. 
A dropout layer with the dropout probability of 0.2 is applied before the final fully-connected layer to the networks with more than 200 layers. 
We also apply DropBlock layers to the convolutional layers at the last two stages of the network. 
As a structured variant of dropout, DropBlock~\cite{ghiasi2018dropblock} randomly masks out local block regions, and is more effective than dropout for specifically regularizing convolutional layers. 

\begin{table*}[t]
\begin{center}
\hspace{-3em}
 \begin{minipage}{0.35\textwidth}%
    \centering
	\subfloat
	{
	\scalebox{0.95}
	    {
	        \begin{tabular} {l | c | c | c  }
            \toprule[1pt]
            & {\#P} & {GFLOPs}  & {acc(\%)} \\
            \cline{1-4}
           ResNetD-50~\cite{he2019bag} & 25.6M & 4.34 & 78.31\\
           + mixup & 25.6M & 4.34 & 79.15\\
           + autoaug & 25.6M & 4.34 & 79.41 \\
            ResNeSt-50-fast & 27.5M & 4.34 & 80.64 \\
            ResNeSt-50 & 27.5M & 5.39 & 81.13 \\
            \bottomrule[1pt]
          \end{tabular}
	    }
	}
 \end{minipage}
 \hspace{3em}
 \begin{minipage}{0.35\textwidth}%
	\centering
	\subfloat
	{
	\scalebox{0.95}
        {
          \begin{tabular} {l | c | c | c | c }
            \toprule[1pt]
            {Variant} & {\#P} & {GFLOPs} & img/sec & {acc(\%)} \\
            \cline{1-5}
            {0s1x64d } & 25.6M & 4.34 & 688.2 & 79.41 \\
            {1s1x64d } & 26.3M & 4.34 & 617.6 & 80.35 \\
            {2s1x64d } & 27.5M & 4.34 & 533.0 & 80.64 \\
            {4s1x64d } & 31.9M & 4.35 & 458.3 & 80.90 \\
            {2s2x40d } & 26.9M & 4.38 & 481.8 & 81.00  \\  
            \bottomrule[1pt]
          \end{tabular}
        }
    }
\end{minipage}
\end{center}
\caption{Ablation study for ImageNet image classification. (Left) breakdown of improvements. (Right) {\it radix vs. cardinality} under ResNeSt-fast setting. 
For example {\it 2s2x40d} denotes radix$=$2, cardinality$=$2 and width$=$40. 
Note that even radix$=$1 does not degrade any existing approach (see Equation~\ref{eq:radix}). 
}
\label{tab:ablation}
\end{table*}

Finally, we also apply weight decay (i.e.\ L2 regularization) which additionally helps stabilize training. 
We only apply weight decay to the weights of convolutional and fully connected layers~\cite{goyal2017accurate,he2019bag}.

\section{Image Classification Results}
\label{sec:exp}

Our first experiments study the image classification performance of ResNeSt on the ImageNet 2012 dataset~\cite{imagenet} with 1.28M training images and 50K validation images (from 1000 different classes). As is standard, networks are trained on the training set and we report their top-1 accuracy on the validation set.

\subsection{Implementation Details}

We use data sharding for distributed training on ImageNet, evenly partitioning the data across GPUs. 
At each training iteration, a mini-batch of training data is sampled from the corresponding shard (without replacement). 
We apply the transformations from the learned Auto Augmentation policy to each individual image. 
Then we further apply standard transformations including: random size crop, random horizontal flip, color jittering, and changing the lighting. 
Finally, the image data are RGB-normalized via mean/standard-deviation rescaling. 
For mixup training, we simply mix each sample from the current mini-batch with its reversed order sample~\cite{he2019bag}. 
Batch Normalization~\cite{ioffe2015batch} is used after each convolutional layer before  ReLU activation~\cite{nair2010rectified}. Network weights are initialized using Kaiming Initialization~\cite{he2015delving}. A drop layer is inserted before the final classification layer with dropout ratio $= 0.2$. Training is done for 270 epochs with a weight decay of 0.0001 and momentum of 0.9, using a cosine learning rate schedule with the first 5 epochs reserved for warm-up. 
We use a mini-batch of size 8192 for ResNeSt-50, 4096 for ResNeSt~101, and 2048 for ResNeSt-\{200, 269\}. 
For evaluation, we first resize each image to 1/0.875 of the crop size along the short edge and apply a center crop. Our code implementation for ImageNet training uses GluonCV~\cite{guo2020gluoncv} with MXNet~\cite{chen2015mxnet}.



\subsection{Ablation Study}


ResNeSt is based on the ResNet-D model~\cite{he2019bag}. 
Mixup training improves the accuracy of ResNetD-50  from 78.31\% to 79.15\%. 
Auto augmentation further improves the accuracy by 0.26\%. 
When employing our Split-Attention block to form a \emph{ResNeSt-50-fast} model, accuracy is further boosted to 80.64\%. 
In this ResNeSt-fast setting, the effective average downsampling is applied prior to the $3\times3$ convolution to avoid introducing extra computational costs in the model. 
With the downsampling operation moved after the convolutional layer, ResNeSt-50 achieves 81.13\% accuracy.

\noindent{\bf Radix vs.\ Cardinality. } 
We conduct an ablation study on 
ResNeSt-variants with different radix/cardinality. In each variant, we adjust the network's width appropriately so that its overall computational cost remains similar to the ResNet variants. 
The results are shown in Table~\ref{tab:ablation}, where $s$ denotes the radix, $x$ the cardinality, and $d$ the network width (0$s$ represents the use of a standard residual block as in ResNet-D~\cite{he2019bag}). 
We empirically find that increasing the radix from 0 to 4 continuously improves the top-1 accuracy, while also increasing latency and memory usage.
Although we expect further accuracy improvements with even greater radix/cardinality, 
we employ Split-Attention with the 2$s$1$x$64$d$ setting in subsequent experiments, to ensure these blocks scale to deeper networks with a good trade-off between speed, accuracy and memory usage.

\subsection{Comparing against the State-of-the-Art}

\begin{table}[t]
	\small
	\renewcommand\arraystretch{1.1}
	\newcommand{\tabincell}[2]{\begin{tabular}{@{}#1@{}}#2\end{tabular}}
	\begin{center}
	\begin{minipage}{\linewidth}
	\centering
	    \resizebox{0.9\textwidth}{!}{
		\begin{tabular}{l|c|c|c|c}
			\toprule[1pt]
			& \#P 
			& crop & img/sec & acc(\%) \\
            \cline{1-5}
			ResNeSt-101(\TextRed{\small ours}) & 48M 
			& 256 & \best 291.3 & \best 83.0 \\
			EfficientNet-B4~\cite{efficientnet} & 19M 
			& 380 &  149.3 &  83.0 \\
			SENet-154~\cite{hu2017squeeze} & 146M 
			& 320 & 133.8 & 82.7 \\
			NASNet-A~\cite{zoph2018learning} & 89M  & 331 & 103.3 
			& 82.7 \\
			AmoebaNet-A~\cite{amoebanet} & 87M 
			& 299 & - & 82.8 \\
			\hdashline
			ResNeSt-200 (\TextRed{\small ours}) & 70M 
			& 320 & \best 105.3 & \best 83.9 \\
			EfficientNet-B5~\cite{efficientnet} & 30M 
			& 456 &  84.3 &  83.7 \\
			AmoebaNet-C~\cite{amoebanet} & 155M 
			& 299 & - & 83.5 \\
			\hdashline
            ResNeSt-269 (\TextRed{\small ours}) & 111M 
            & 416 & \best 51.2 & \best 84.5 \\
			GPipe & 557M 
			& - & - & 84.3 \\
			EfficientNet-B7~\cite{efficientnet} & 66M 
			& 600 & 34.9 & 84.4 \\
			\bottomrule[1pt]
		\end{tabular}
	    }
	\end{minipage}
	
	\end{center}
	\vspace{+1pt}
	\caption{
	Accuracy vs. Throughput for SoTA CNN models on ImageNet. 
	Our ResNeSt model displays the best trade-off.  
	Average Inference latency is measured on a NVIDIA V100 GPU using the original code implementation of each model with a mini-batch of size 16.
	}
	\label{tab:cls_sota}
	\vspace{-1em}
\end{table}


To compare with CNN models trained using different crop size settings, we increase the training crop size for deeper models. 
We use a crop size of $256 \times 256$ for ResNeSt-200 and $320\times 320$ for ResNeSt-269. Bicubic upsampling strategy is employed for input-size greater than 256. 
The results are shown in Table~\ref{tab:cls_sota}, where we compare the inference speed in addition to the number of parameters. We find that despite its advantage in parameters with accuracy trade-off, the widely used depth-wise convolution is not optimized for inference speed. 
In this benchmark, all inference speeds are measured using a mini-batch of 16 using the implementation~\cite{efficientnet_github} from the original author on a single NVIDIA V100 GPU. 
The proposed ResNeSt has better accuracy and latency trade-off than models found via neural architecture search.

\section{Transfer Learning Results}
\label{sec:transferexp}

\subsection{Object Detection}

We report our detection result on MS-COCO\cite{lin2014microsoft} in Table~\ref{tab:mscoco_det}. All models are trained on COCO-2017 training set with 118k images, and evaluated on COCO-2017 validation set with 5k images (aka. minival) using the standard COCO AP metric of single scale. 
We train all models with FPN\cite{lin2017feature}, synchronized batch normalization~\cite{Zhang_2018_CVPR} and image scale augmentation (short size of a image is picked randomly from 640 to 800). 1x learning rate schedule is used. 
We conduct Faster-RCNNs and Cascade-RCNNs experiments using Detectron2\cite{wu2019detectron2}. 
For comparison, we simply replaced the vanilla ResNet backbones with our ResNeSt, while using the default settings for the hyper-parameters and detection heads \cite{gluoncvnlp2020, wu2019detectron2}.

Compared to the baselines using standard ResNet, Our backbone is able to boost mean average precision by around 3\% on both Faster-RCNNs and Cascade-RCNNs. 
The result demonstrates our backbone has good generalization ability and can be easily transferred to the downstream task. 
Notably, our ResNeSt50 outperforms ResNet101 on both Faster-RCNN and Cascade-RCNN detection models, using significantly fewer parameters. 
Detailed results in Table~\ref{tab:mscoco_det}. 
We evaluate our Cascade-RCNN with ResNeSt101 deformable, that is trained using 1x learning rate schedule on COCO test-dev set as well. It yields a box mAP of 49.2 using single scale inference.


\begin{table}[t]
\begin{center}
\resizebox{\linewidth}{!}
  {
  \begin{tabular} {l | l | c | c }
    \toprule[1pt]
    & {\bf Method} & {\bf Backbone} & {\bf mAP\%} 
    \\
    \cline{2-4}
    \cline{2-4}
    \multirow{5}{*}{ \rotatebox{90}{Prior Work}} & \multirow{3}{*}{Faster-RCNN~\cite{ren2015faster}} & ResNet101~\cite{he2017mask} & 37.3 \\
    &  & ResNeXt101~\cite{xie2016aggregated, mmdetection} & 40.1 \\
    &  & SE-ResNet101~\cite{hu2017squeeze} & 41.9 \\ 
    \cdashline{2-4}
    & Faster-RCNN+DCN~\cite{dai2017deformable} & ResNet101~\cite{mmdetection} & 42.1 \\
    & Cascade-RCNN~\cite{cai2018cascade} & ResNet101 & 42.8 \\
	\hdashline
    \multirow{9}{*}{ \rotatebox{90}{Our Results}}& \multirow{4}{*}{Faster-RCNN~\cite{ren2015faster}} & ResNet50~\cite{wu2019detectron2} & 39.25 \\
    & & ResNet101~\cite{wu2019detectron2} &  41.37 \\
    & & ResNeSt50 (\TextRed{\small ours}) & \secbest 42.33 \\
    & & ResNeSt101 (\TextRed{\small ours}) & \best 44.72  \\
    \cdashline{2-4}
    & \multirow{4}{*}{Cascade-RCNN~\cite{cai2018cascade}} & ResNet50~\cite{wu2019detectron2} &  42.52 \\
    & & ResNet101~\cite{wu2019detectron2} &  44.03 \\
    & & ResNeSt50 (\TextRed{\small ours}) &  \secbest 45.41 \\
    & & ResNeSt101 (\TextRed{\small ours}) & \best 47.50 \\ 
    \cdashline{2-4} 
    & Cascade-RCNN~\cite{cai2018cascade} & ResNeSt200 (\TextRed{\small ours}) & 49.03 \\ 
    \bottomrule[1pt]
  \end{tabular}
  }
\end{center}
\caption{Object detection results on the MS-COCO validation set. Both Faster-RCNN and Cascade-RCNN are significantly improved by our ResNeSt backbone. }
\vspace{-2em}
\label{tab:mscoco_det}
\end{table}

\subsection{Instance Segmentation}

To explore the generalization ability of our novel backbone, we also apply it to instance segmentation tasks. 
Besides the bounding box and category probability, instance segmentation also predicts object masks, for which a more accurate dense image representation is desirable. 

We evaluate the Mask-RCNN~\cite{he2017mask} and Cascade-Mask-RCNN~\cite{cai2018cascade} models with ResNeSt-50 and ResNeSt-101 as their backbones. All models are trained along with FPN~\cite{lin2017feature} and synchronized batch normalization. For data augmentation, input images' shorter side are randomly scaled to one of (640, 672, 704, 736, 768, 800). To fairly compare it with other methods, 1x learning rate schedule policy is applied, and other hyper-parameters remain the same. We re-train the baseline with the same setting described above, but with the standard ResNet. All our experiments are trained on COCO-2017 dataset and using Detectron2\cite{wu2019detectron2}. 
For the baseline experiments, the backbone we used by default is the MSRA version of ResNet, having stride-2 on the 1x1 conv layer.
Both bounding box and mask mAP are reported on COCO-2017 validation dataset.

As shown in Table~\ref{tab:instance_segmentation}, our new backbone achieves better performance. For Mask-RCNN, ResNeSt50 outperforms the baseline with a gain of 2.85\%/2.09\% for box/mask performance, and ResNeSt101 exhibits even better improvement of 4.03\%/3.14\%. For Cascade-Mask-RCNN, the gains produced by switching to ResNeSt50 or ResNeSt101 are 3.13\%/2.36\% or 3.51\%/3.04\%, respectively. This suggests a model will be better if it consists of more Split-Attention modules. As observed in the detection results, the mAP of our ResNeSt50 exceeds the result of the standard ResNet101 backbone, which indicates a higher capacity of the small model with our proposed module. 
Finally, we also train a Cascade-Mask-RCNN with ResNeSt101-deformable using a 1x learning rate schedule. We evaluate it on the COCO test-dev set, yielding 50.0 box mAP, and 43.1 mask mAP respectively. 
Additional experiments under different settings are included in the supplementary material.


\begin{table}[t]
\begin{center}
\resizebox{\linewidth}{!}
  {
  \begin{tabular} {l | l | c | c c }
    \toprule[1pt]
    \multirow{4}{*}{ \rotatebox{90}{Prior Work } } & {\bf Method} & {\bf Backbone} & {\bf box mAP\%} & 
    {\bf mask mAP\%} 
    \\
    \cline{2-5}
    \cline{2-5}
     & DCV-V2~\cite{zhu2019deformable} & ResNet50 & 42.7 & 37.0 \\
    & HTC~\cite{chen2019hybrid} & ResNet50 & 43.2 &  38.0 \\
    & Mask-RCNN~\cite{he2017mask} & ResNet101~\cite{mmdetection} & 39.9 & 36.1 \\
    & Cascade-RCNN~\cite{cai2019cascade} & ResNet101 & 44.8 & 38.0 \\
	\hdashline
    \multirow{8}{*}{ \rotatebox{90}{Our Results}}& \multirow{4}{*}{Mask-RCNN~\cite{he2017mask}} & ResNet50~\cite{wu2019detectron2} & 39.97 & 36.05     \\
    & & ResNet101~\cite{wu2019detectron2} & 41.78  & 37.51  \\
    & & ResNeSt50 (\TextRed{\small ours}) & \secbest 42.81 & \secbest 38.14 \\
    & & ResNeSt101 (\TextRed{\small ours}) & \best 45.75 & \best 40.65 \\
	\cdashline{2-5}
    & \multirow{4}{*}{Cascade-RCNN~\cite{cai2018cascade}} & ResNet50~\cite{wu2019detectron2} & 43.06 & 37.19 \\
    & & ResNet101~\cite{wu2019detectron2} & 44.79 &  38.52 \\
    & & ResNeSt50 (\TextRed{\small ours}) & \secbest 46.19 & \secbest 39.55 \\
    & & ResNeSt101 (\TextRed{\small ours}) & \best 48.30 & \best 41.56 \\
    \bottomrule[1pt]
  \end{tabular}
  }
  \end{center}
\caption{Instance Segmentation results on the MS-COCO validation set. 
Both Mask-RCNN and Cascade-RCNN models are improved by our ResNeSt backbone. Models with our ResNeSt-101 outperform all prior work using ResNet-101. 
}
\label{tab:instance_segmentation}
\vspace{-1em}
\end{table}

\subsection{Semantic Segmentation}

In transfer learning for semantic segmentation, we use the GluonCV~\cite{guo2020gluoncv} implementation of DeepLabV3~\cite{chen2017rethinking} as a baseline approach. 
Here a dilated network strategy~\cite{chen2016Deeplab,yu2015multi} is applied to the backbone network, resulting in a stride-8 model. 
Synchronized Batch Normalization~\cite{Zhang_2018_CVPR} is used during training, along with a polynomial-like learning rate schedule (with initial learning rate $= 0.1$). 
For evaluation, the network prediction logits are upsampled 8 times to calculate the per-pixel cross entropy loss against the ground truth labels. 
We use multi-scale evaluation with flipping~\cite{zhao2016pyramid,Zhang_2018_CVPR,Zhu2019VPLR}.

We first consider the Cityscapes~\cite{cordts2016cityscapes} dataset, which consists of 5K high-quality labeled images.  We train each model on 2,975 images from the training set and report its mIoU on 500 validation images. Following  prior work, we only consider 19 object/stuff categories in this benchmark. We have not used any coarse labeled images or any extra data in this benchmark. 
Our ResNeSt backbone boosts the mIoU achieved by DeepLabV3 models by around 1\% while maintaining a similar overall model complexity. Notably, the DeepLabV3 model using our ResNeSt-50 backbone already achieves better performance than DeepLabV3 with a much larger ResNet-101 backbone.

\begin{table}[t]
	\centering
	\subfloat
	{
	\scalebox{0.8}
      {
      \begin{tabular} {l | l | c | c  c }
        \toprule[1pt]
        & {\bf Method} & {\bf Backbone} & {\bf pixAcc\%} & 
        {\bf mIoU\%} \\
        \hline \hline
        \multirow{6}{*}{ \rotatebox{90}{Prior Work}} & UperNet \cite{xiao2018unified}  & ResNet101 & 81.01 & 
        42.66 \\
        & PSPNet \cite{zhao2016pyramid} & ResNet101 & 81.39 &  
        43.29 \\
        & EncNet \cite{Zhang_2018_CVPR} &  ResNet101 & 81.69 &  
         44.65 \\
    	& CFNet \cite{zhang2019co} & ResNet101 & 81.57 & 
    	44.89 \\
    	& OCNet \cite{yuan2019object} & ResNet101 & - & 45.45 \\
    	& ACNet \cite{fu2019adaptive} & ResNet101 &  81.96 &  45.90 \\
    	\hdashline
        \multirow{5 }{*}{ \rotatebox{90}{Ours}} &\multirow{5}{*}{DeeplabV3~\cite{chen2017rethinking}} & ResNet50~\cite{guo2020gluoncv} & 80.39 & 42.1 \\
        & & ResNet101~\cite{guo2020gluoncv} & 81.11 & 44.14 \\
        & & ResNeSt-50 (\TextRed{\small ours}) & \secbest 81.17 & \secbest 45.12 \\
        & & ResNeSt-101 (\TextRed{\small ours}) & \best 82.07 & \best 46.91  \\
        \cdashline{3-5}
        & & ResNeSt-200 (\TextRed{\small ours}) & 82.45 &  48.36  \\
        \bottomrule[1pt]
      \end{tabular}
      }
  }

 \caption{Semantic segmentation results on validation set of: ADE20K. 
}
\end{table}

\begin{table}[t]
	\centering
	\subfloat
	{
		\scalebox{0.8} {
		\begin{tabular} {l | l | c | c }
            \toprule[1pt]
            & {\bf Method} & {\bf Backbone}  & 
            {\bf mIoU\%} \\
            \hline \hline
            \multirow{6}{*}{ \rotatebox{90}{Prior Work}} & DANet \cite{Fu2018DANet} & ResNet101 & 77.6 \\
            & PSANet \cite{zhao2018psanet} & ResNet101 & 77.9 \\
            & PSPNet \cite{zhao2016pyramid} & ResNet101 & 78.4 \\
            & AAF \cite{Ke2018AAF} & ResNet101 & 79.2 \\
            & DeeplabV3~\cite{chen2017rethinking} & ResNet101 & 79.3 \\
        	& OCNet \cite{yuan2019object} & ResNet101 & 80.1 \\
        	\hdashline
            \multirow{4}{*}{ \rotatebox{90}{Ours}} & \multirow{4}{*}{DeeplabV3~\cite{chen2017rethinking}} &  ResNet50~\cite{guo2020gluoncv} & 78.72 \\
            & & ResNet101~\cite{guo2020gluoncv} & 79.42 \\
            & & ResNeSt-50 (\TextRed{\small ours}) & \secbest 79.87 \\
            & & ResNeSt-101 (\TextRed{\small ours}) & \best 80.42  \\
            \cdashline{3-4}
            & & ResNeSt-200 (\TextRed{\small ours}) &  82.7  \\
            \bottomrule[1pt]
        \end{tabular}
	    }
	}

\caption{Semantic segmentation results on validation set of Citscapes. Models are trained without coarse labels or extra data.  
}
\vspace{-1em}
\label{tab:ade20k}
\end{table}

ADE20K~\cite{zhou2017scene} is a large scene parsing dataset with 150 object and stuff classes containing 20K training, 2K validation, and 3K test images. 
All networks are trained on the training set for 120 epochs and evaluated on the validation set. Table~\ref{tab:ade20k} shows the resulting pixel accuracy (pixAcc) and mean intersection-of-union (mIoU). 
The performance of the DeepLabV3 models are dramatically improved by employing our ResNeSt backbone. 
Analogous to previous results, the DeepLabv3 model using our ResNeSt-50 backbone already outperforms DeepLabv3 using a deeper ResNet-101 backbone. DeepLabV3 with a ResNeSt-101 backbone achieves 82.07\% pixAcc and 46.91\% mIoU, which to our knowledge, is the best single model that has been presented for ADE20K.

\section{Conclusion}

This work proposes the ResNeSt architecture that leverages the channel-wise attention with multi-path representation into a single unified Split-Attention block. 
The model universally improves the learned feature representations to boost performance across image classification, object detection, instance segmentation and semantic segmentation. 
Our Split-Attention block is easy to work with (i.e., drop-in replacement of a standard residual block), computationally efficient (i.e., 32\% less latency than EfficientNet-B7 but with better accuracy), and transfers well. We believe ResNeSt can have an impact across multiple vision tasks, as it has already been adopted by multiple winning entries in 2020 COCO-LVIS challenge and 2020 DAVIS-VOS chanllenge.

{\small
\bibliographystyle{ieee_fullname}
\bibliography{egbib}
}

\clearpage
\appendix
\section*{Appendix}

\begin{table}[t]
\begin{center}
\resizebox{\linewidth}{!}
  {
  \begin{tabular} { l | c | c | c}
    \toprule[1pt]
    {Method} & {Backbone}  & {OKS AP\% w/o flip}  & {OKS AP\% w/ flip}  \\
    \hline
    \multirow{4}{*}{SimplePose~\cite{xiao2018simple}} 
    & ResNet50~\cite{guo2020gluoncv}  & 71.0/91.2/78.6 &  72.2/92.2/79.9 \\ 
    & ResNet101~\cite{guo2020gluoncv} & 72.6/91.3/80.8 & 73.6/92.3/81.1 \\ 
    & ResNeSt50 (\TextRed{\small ours}) & 72.3/92.3/80.0 & 73.4/92.4/81.2  \\ 
    & ResNeSt101 (\TextRed{\small ours}) & 73.6/92.3/80.9 & 74.6/92.4/82.1 \\ 
    \bottomrule[1pt]
  \end{tabular}
  }
\end{center}
\caption{Pose estimation results on MS-COCO dataset in terms of OKS AP.}
\label{tab:pose}
\end{table}

\begin{table}[t]
\begin{center}
\resizebox{\linewidth}{!}
  {
  \begin{tabular} {l | l | c | c | c }
    \toprule[1pt]
    & {\bf Method} & {\bf Backbone} & Deformable & {\bf mAP\%} 
    \\
    \cline{2-5}
    \cline{2-5}
    \multirow{4}{*}{ \rotatebox{90}{Prior Work}} & \multirow{2}{*}{DCNv2~\cite{zhu2019deformable}} & ResNet101~\cite{he2015deep} & v2 & 44.8 \\
    &  & ResNeXt101~\cite{xie2016aggregated} & v2 & 45.3 \\
    \cdashline{2-5}
    & \multirow{2}{*}{TridentNet~\cite{dai2017deformable}} & ResNet101~\cite{he2015deep} & v1& 46.8 \\
    &  & ResNet101*~\cite{dai2017deformable} & v1 & 48.4 \\
    \cdashline{2-5}
    & SNIPER~\cite{singh2018sniper} & ResNet101*~\cite{dai2017deformable} & v1 & 46.1 \\
    \cdashline{2-5}
    & Cascade-RCNN~\cite{cai2019cascade} & ResNet101~\cite{he2015deep} & n/a & 42.8 \\
    \hline
    & \multirow{1}{*}{Cascade-RCNN~\cite{cai2019cascade}} & ResNeSt101 (\TextRed{\small ours}) & v2 & \best 49.2 \\ 
    \bottomrule[1pt]
  \end{tabular}
  }
\end{center}
\caption{Object detection results on the MS-COCO test-dev set. The single model of Cascade-RCNN with ResNeSt backbone using deformable convolution~\cite{dai2017deformable} achieves 49.2\% mAP, which outperforms all previous methods. (* means using multi-scale evaluation.)}
\label{tab:mscoco_det}
\end{table}

\subsection{Pose Estimation}

We investigate the effect of backbone on pose estimation task. The baseline model is SimplePose~\cite{xiao2018simple} with ResNet50 and ResNet101 implemented in GluonCV~\cite{guo2020gluoncv}. As comparison we replace the backbone with ResNeSt50 and ResNeSt101 respectively while keeping other settings unchanged. The input image size is fixed to 256x192 for all runs. We use Adam optimizer with batch size 32 and initial learning rate 0.001 with no weight decay. The learning rate is divided by 10 at the 90th and 120th epoch. The experiments are conducted on COCO Keypoints dataset, and we report the OKS AP for results without and with flip test. Flip test first makes prediction on both original and horizontally flipped images, and then averages the predicted keypoint coordinates as the final output.

From Table ~\ref{tab:pose}, we see that models backboned with ResNeSt50/ResNeSt101 significantly outperform their ResNet counterparts. Besides, with ResNeSt50 backbone the model achieves performance similar with ResNet101 backbone.

\subsection{Object Detection and Instance Segmentation}

For object detection, we add deformable convolution to our Cascade-RCNN model with ResNeSt-101 backbone and train the model on the MS-COCO training set for 1x schedule. 
The resulting model achieves 49.2\% mAP on COCO test-dev set, which surpass all previous methods including these employing multi-scale evaluation. 
Detailed results are shown in Table~\ref{tab:mscoco_det}.

\begin{table}[t]
\begin{center}
\resizebox{\linewidth}{!}
  {
  \begin{tabular} {l | l | c | c | c | c }
    \toprule[1pt]
    & {\bf Method} & {\bf Backbone} & Deformable & {\bf box mAP\%} & {\bf mask mAP\%} 
    \\
    \cline{2-6}
    \multirow{1}{*}{\rotatebox{90}{Prior Work}} & \multirow{2}{*}{DCNv2~\cite{zhu2019deformable}} & ResNet101~\cite{he2015deep} & v2 & 45.8 & 39.7 \\
    &  & ResNeXt101~\cite{xie2016aggregated} & v2 & 46.7 & 40.5 \\
    \cdashline{2-6}
    & SNIPER~\cite{singh2018sniper} & ResNet101*~\cite{dai2017deformable} & v1 & 47.1 & 41.3 \\
    \cdashline{2-6}
    & Cascade-RCNN~\cite{cai2019cascade} & ResNext101~\cite{xie2016aggregated} & n/a & 45.8 & 38.6 \\
    \hline
    & \multirow{1}{*}{Cascade-RCNN~\cite{cai2019cascade}} & ResNeSt101 (\TextRed{\small ours}) & v2 & \best 50.0 & \best 43.0 \\ 
    \bottomrule[1pt]
  \end{tabular}
  }
\end{center}
\caption{Instance Segmentation results on the MS-COCO test-dev set. * denote multi-scale inference.}
\label{tab:mscoco_det}
\end{table}

\begin{table}
\begin{center}
\resizebox{\linewidth}{!}
  {
  \begin{tabular} {l | c | c | c | c c }
    \toprule[1pt]
    {\bf Method} & {\bf lr schedule} & {\bf SyncBN} & {\bf Backbone} & {\bf box mAP\%} & 
    {\bf mask mAP\%} 
    \\
	\hline
	
    \multirow{12}{*}{Mask-RCNN~\cite{he2017mask}} & \multirow{8}{*}{1$\times$}  & & ResNet50~\cite{wu2019detectron2} & 38.60 & 35.20       \\
    & & & ResNet101~\cite{wu2019detectron2} & 40.79  & 36.93   \\
    & & &  ResNeSt50 (\TextRed{\small ours}) & \secbest 40.85  & \secbest 36.99  \\
    & & & ResNeSt101 (\TextRed{\small ours}) & \best 43.98  & \best 39.33  \\
    
    \cdashline{3-6}
    & & \multirow{4}{*}{\checkmark} & ResNet50~\cite{wu2019detectron2} & 39.97  & 36.05  \\ 
    & & & ResNet101~\cite{wu2019detectron2} & 41.78 & 37.51  \\
    & & & ResNeSt50 (\TextRed{\small ours}) & \secbest 42.81  & \secbest 38.14    \\
    & & & ResNeSt101 (\TextRed{\small ours}) & \best 45.75  & \best 40.65  \\
    
    \cline{2-6}
    
    & \multirow{4}{*}{3$\times$} & &  ResNet50~\cite{wu2019detectron2} & 41.00 & 37.20 \\ 
    & & & ResNet101~\cite{wu2019detectron2} & 42.90 & 38.60 \\
    & & & ResNeSt50 (\TextRed{\small ours}) & \secbest 43.32 & \secbest 38.91   \\
    & & & ResNeSt101 (\TextRed{\small ours}) & \best 45.37 & \best 40.56 \\

	\hline

    \multirow{12}{*}{Cascade-RCNN~\cite{cai2018cascade}} & \multirow{8}{*}{1$\times$} & & ResNet50~\cite{wu2019detectron2} & 42.10 & 36.40  \\
    & & & ResNet101~\cite{wu2019detectron2} & 44.00  & 38.08   \\
    & & & ResNeSt50 (\TextRed{\small ours}) & \secbest 44.56  & \secbest 38.27  \\
    & & & ResNeSt101 (\TextRed{\small ours}) & \best 46.86  & \best 40.23  \\
    
    \cdashline{3-6}
    & &\multirow{4}{*}{\checkmark} & ResNet50~\cite{wu2019detectron2} & 43.06  & 37.19  \\ 
    & & & ResNet101~\cite{wu2019detectron2} & 44.79 & 38.52  \\
    & & & ResNeSt50 (\TextRed{\small ours}) & \secbest 46.19  & \secbest 39.55    \\
    & & & ResNeSt101 (\TextRed{\small ours}) & \best 48.30  & \best 41.56  \\

    \cline{2-6}

    & \multirow{4}{*}{3$\times$} & &  ResNet50~\cite{wu2019detectron2} & 44.3 & 38.5 \\ 
    & & & ResNet101~\cite{wu2019detectron2} & 45.57 & 39.54 \\
    & & & ResNeSt50 (\TextRed{\small ours}) & \secbest 46.39  & \secbest 39.99   \\
    & & & ResNeSt101 (\TextRed{\small ours}) & \best 47.70 & \best 41.16   \\

    \bottomrule[1pt]
  \end{tabular}
  }
\end{center}
\caption{Instance Segmentation results on the MS-COCO validation set. Comparing models trained w/ and w/o SyncBN, and using 1$\times$ and 3$\times$ learning rate schedules. 
}
\label{tab:supp_instance_segmentation}
\end{table}

\begin{table}
\begin{center}
\resizebox{0.7\linewidth}{!}
  {
  \begin{tabular} {l |  c | c c }
    \toprule[1pt]
    
    {\bf Method} & {\bf Deformable~\cite{zhu2019deformable} (v2)} & {\bf box mAP\%} & {\bf mask mAP\%} \\
    \hline
    
    \multirow{2}{*}{Cascade-RCNN~\cite{cai2018cascade}} &  & 48.30  & 41.56  \\
   
    \cdashline{2-4}
    & \checkmark & \best 49.39 & \best 42.56   \\
    
    \bottomrule[1pt]
  \end{tabular}
  }
  \end{center}
\caption{
The results of Cascade-Mask-RCNN on COCO val set. The ResNeSt-101 is applied with and without deformable convolution v2\cite{zhu2019deformable}. It shows that our split-attention module is compatible with other existing modules.}
\label{tab:supp_instance_segmentation_dcn}
\end{table}

We include more results of instance segmentation, shown in Table~\ref{tab:supp_instance_segmentation}, from the models trained with 1x/3x learning rate schedules and with/without SyncBN. All of resutls are reported on COCO val dataset. For both 50/101-layer settings, our ResNeSt backbones still outperform the corresponding baselines with different lr schedules. Same as the Table. 6 in the main text, our ResNeSt50 also exceeds the result of the standard ResNet101.

We also evaluate our ResNeSt with and without deformable convolution v2\cite{zhu2019deformable}. With its help, we are able to obtain a higher performance, shown in Table~\ref{tab:supp_instance_segmentation_dcn}. It indicates our designed module is compatible with deformable convolution.\\





\end{document}